\documentclass[10pt,twocolumn,letterpaper]{article}

\usepackage[pagenumbers]{iccv} % To force page numbers, e.g. for an arXiv version

\definecolor{iccvblue}{rgb}{0.21,0.49,0.74}
\usepackage[pagebackref,breaklinks,colorlinks,allcolors=iccvblue]{hyperref}

\def\paperID{*****} % *** Enter the Paper ID here
\def\confName{ICCV}
\def\confYear{2025}

\title{2D Instance Editing in 3D Space}

\author{
\textbf{Yuhuan Xie}\textsuperscript{1}
\ 
\textbf{Aoxuan Pan}\textsuperscript{2}
\ 
\textbf{Ming-Xian Lin}\textsuperscript{1}
\ 
\textbf{Wei Huang}\textsuperscript{1}
\ 
\textbf{Yi-Hua Huang}\textsuperscript{1$\dagger$}
\ 
\textbf{Xiaojuan Qi}\textsuperscript{1$\dagger$} \\
\textsuperscript{1} The University of Hong Kong
\quad
\textsuperscript{2} Independent
}

\begin{document}

\twocolumn[{%
\renewcommand\twocolumn[1][]{#1}%
\maketitle

\begin{center}
    \centering
    \captionsetup{type=figure}
    \includegraphics[width=1\linewidth, trim=0 0 0 0, clip]{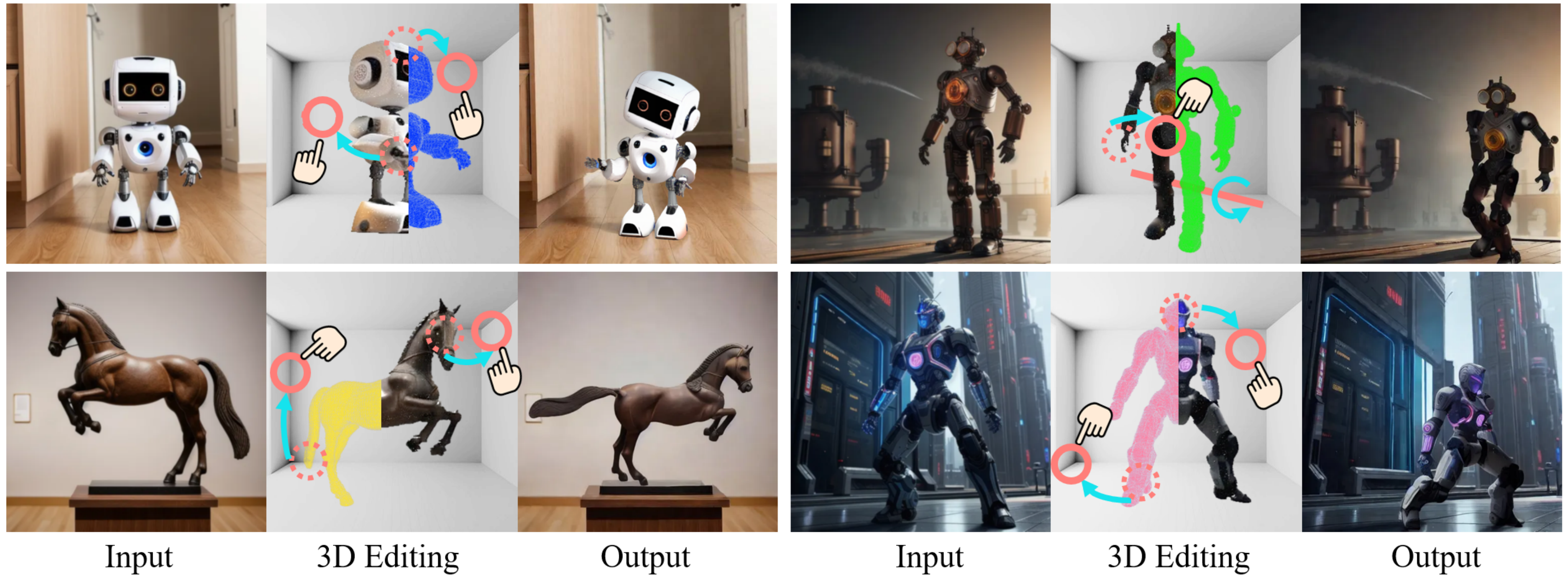}
    \vspace{-6mm}
    \captionof{figure}{
    Given 2D instances in images, we propose a method to lift them into 3D, enabling edits under the local rigidity assumption. This approach facilitates large view changes, non-rigid deformations, and interactive, real-time operations for more efficient user control.
    }
    \vspace{-1mm}
\label{fig:teaser}
\end{center}
}]

\if TT\insert\footins{\noindent\footnotesize{
$\dagger$Corresponding Author}}\fi

\begin{abstract}
Generative models have achieved significant progress in advancing 2D image editing, demonstrating exceptional precision and realism. However, they often struggle with consistency and object identity preservation due to their inherent pixel-manipulation nature. To address this limitation, we introduce a novel "2D-3D-2D" framework. Our approach begins by lifting 2D objects into 3D representation, enabling edits within a physically plausible, rigidity-constrained 3D environment. The edited 3D objects are then reprojected and seamlessly inpainted back into the original 2D image. In contrast to existing 2D editing methods, such as DragGAN and DragDiffusion, our method directly manipulates objects in a 3D environment. Extensive experiments highlight that our framework surpasses previous methods in general performance, delivering highly consistent edits while robustly preserving object identity.

\end{abstract}
\section{Introduction}

% The rapid advancement of deep generative models has fundamentally revolutionized digital image editing. Beyond traditional filters and color adjustments, users now have advanced tools to manipulate image content with greater precision and realism, enabling direct edits through prompt commands~\cite{brooks2023instructpix2pix,zhang2023magicbrush} or reference-based guidance~\cite{yang2023paint,kim2023reference}. The field of image editing has witnessed significant growth, fueled by increasingly powerful generative models.

The rapid advancements in deep generative models have fundamentally transformed digital image editing, moving beyond traditional techniques such as filters and color adjustments. These models now provide users with sophisticated tools to manipulate image content with unprecedented precision and realism, enabling direct edits through text-based prompts~\cite{brooks2023instructpix2pix,zhang2023magicbrush} or reference-based guidance~\cite{yang2023paint,kim2023reference}. This evolution has driven significant progress in the field of image editing, propelled by the growing capabilities of generative models. Despite their advancements, most existing approaches rely on high-level semantics derived from language or images as control signals, which often lack the granularity required for precise, fine-grained control over image details. Incorporating interactive controls with intermediate editing results offers a promising solution, enabling real-time feedback that facilitates more accurate adjustments and flexible refinements.

% However, most existing approaches rely on high-level semantics derived from language or images as control signals, which often lack the granularity needed for precise, fine-level control over image details. Introducing interactive controls with intermediate editing results can provide users with real-time feedback, facilitating better adjustments and more flexible refinements.

While some recent works, such as DragGAN~\cite{pan2023draggan} and DragDiffusion~\cite{shi2024dragdiffusion}, enable interactive image edits by allowing users to pick and drag control points, these methods are constrained to editing a limited range of categories, largely due to the restricted capabilities of their data-driven generative models. Moreover, both DragGAN and DragDiffusion operate as 2D image models, "manipulating" pixel data rather than addressing the underlying 3D structure of objects. This limitation affects consistency and the preservation of object identity in the edits.
We hold that a more intuitive approach to image editing should involve directly manipulating objects in 3D space like a sculpture. This method aligns more closely with human understanding of the 3D physical world, enabling more efficient and effective editing to achieve desired results with greater control and accuracy.

To achieve this goal, we propose a pipeline consisting of several steps: lifting 2D objects to 3D, editing 3D objects under a physically plausible rigidity assumption, positioning the 3D objects, and inpainting them back into the original image.
First, we employ SAM~\cite{kirillov2023segment} to segment the target object in the image. Next, we leverage the state-of-the-art 3D generation method, TRELLIS~\cite{xiang2025structured}, to produce a 3D Gaussian Splatting (3DGS)~\cite{kerbl3Dgaussians} representation of the object. This approach is chosen for its real-time rendering capabilities and high visual quality.
Using the generated 3DGS, we develop a real-time interactive editing algorithm that deforms the 3DGS under a local rigidity assumption, ensuring structural preservation and physical fidelity. Sparse control points are sampled from the 3DGS point cloud, and neighboring points are connected to construct a topology-aware graph. We then optimize the local rigid energy with editing constraints, which drives the deformation of the dense 3DGS.
Once the 3D object has been edited, it is placed back into the image at the specified 6DoF pose (translation + rotation). Finally, the surrounding region is inpainted to ensure the results are visually harmonious.

Our main contributions can be summarized as:  

\begin{itemize}
    \item We propose a novel and comprehensive framework for 3D-aware object editing from a single 2D image. This framework enables large-scale 3D deformation and placement while preserving object identity far better than 2D-based methods.  
    \item We develop a real-time interactive 3DGS algorithm to achieve this goal. By leveraging a rigidity assumption and sparse graph modeling, our method enables real-time editing of 3DGS with physically plausible effects.  
    \item Through extensive experiments, we demonstrate that our approach produces high-fidelity, 3D-consistent results across a wide variety of objects, significantly outperforming prior methods in both realism and user control. 
\end{itemize}

\begin{figure*}[t]
\begin{center}  
\includegraphics[width=\linewidth]{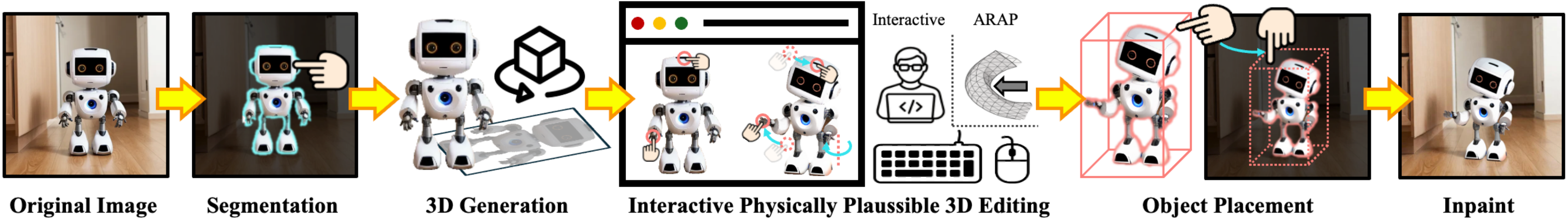}
\vspace{-10mm}
\end{center}
\caption{Our method involves lifting 2D instances into 3D space, editing them under the assumption of local rigidity, and then repositioning them back into the desired locations before inpainting them into the images.}
\label{fig:pipeline}
\vspace{-5mm}
\end{figure*}

\section{Related Work}
\label{sec:related_work}

Our work rethinks 2D image editing by lifting editable object instances into 3D space, enabling geometry-aware manipulation and coherent reintegration within the original scene. We review related works in three interconnected domains: 2D image editing with generative models, lifting 3D geometry from single images, and 3D-aware object editing.

\subsection{2D Image Editing with Generative Models}
\textbf{Classical methods.}
Traditional image editing methods provided algorithmic tools for tasks such as inpainting, resizing, and compositing. Notable examples include Navier-Stokes inpainting~\cite{au2001image}, seam carving~\cite{avidan2023seam}, and Poisson blending~\cite{afifi2015mpb}. While effective in controlled cases, these methods often fail to capture the semantic or geometric structure of complex scenes.

\noindent\textbf{Learning-based image editing.}
The introduction of deep generative models brought greater flexibility and realism to image editing. GAN-based frameworks like Pix2Pix~\cite{henry2021pix2pix}, CycleGAN~\cite{chu2017cyclegan}, and StyleGAN~\cite{karras2019style} enabled tasks such as image-to-image translation and latent space manipulation. These methods support applications like facial control or style transfer, though they may struggle with fine-grained control and even out-of-distribution generalization.
Diffusion models have further advanced editing capabilities, offering improved stability and fidelity. Stable Diffusion~\cite{rombach2022high} and DALL-E 2 allow high-quality synthesis from text prompts, while Prompt-to-Prompt~\cite{hertz2022prompt} and InstructPix2Pix~\cite{brooks2023instructpix2pix} introduce more localized, controllable edits. ControlNet~\cite{zhang2023adding} enhances diffusion-based editing by incorporating spatial conditions such as edges, poses, or depth. Despite their strengths, these models remain fundamentally 2D and lack mechanisms to enforce 3D consistency.

\noindent\textbf{Interactive point-based editing.}
Recent systems such as DragGAN~\cite{pan2023drag} and its diffusion-based extensions~\cite{shi2024dragdiffusion, mou2023dragondiffusion} offer interactive control over image content through point pairs. These tools enable users to specify the movement or deformation of an object, yielding visually striking outcomes. However, the manipulation occurs entirely in the image plane, without explicit modeling of depth. Such limitations make it difficult to maintain physical plausibility, particularly for edits involving large viewpoint changes.

\noindent\textbf{Our perspective.}
Unlike the methods above, our approach lifts object edits into a 3D-aware space. By reconstructing a geometry-consistent representation of the object, we enable edits that align with the underlying structure of the object and preserve coherence during reintegration.

\subsection{Single-View Lifting from 2D to 3D Space}
Estimating 3D structure from a single image has long been a fundamental problem in computer vision. Early deep learning approaches aimed to recover explicit shapes, such as voxel grids or surface meshes~\cite{choy20163d}. More recent work has shifted toward implicit neural representations, which tend to generalize better and produce smoother results.
The introduction of Neural Radiance Fields (NeRF)~\cite{mildenhall2021nerf} marked a turning point for view synthesis, although it originally required multiple views of a scene. Follow-up work gradually adapted NeRF to settings with fewer or even single images. For instance, Zero-1-to-3~\cite{liu2023zero123} uses large-scale diffusion models to generate novel views from just one input image, capturing a strong sense of object geometry. Meanwhile, 3D Gaussian Splatting~\cite{kerbl20233d} provides an efficient and high-fidelity way to represent and render scenes using explicit point-based primitives.
These developments have made it feasible to lift a single image into a 3D-aware representation. In our case, this capability is not used for rendering new views, but instead forms the basis for editable object manipulation and seamless re-integration within the original scene.

\subsection{3D-Aware Object Editing}
3D-aware editing aims to modify image content in ways that are consistent with the underlying geometry of the scene. One line of research focuses on 3D-aware generative models, such as EG3D~\cite{chan2022eg3d}, which offer explicit control over object pose and appearance. These methods are typically trained on narrow object categories like faces or cars, and their ability to generalize to arbitrary scenes is limited. Another direction explores scene-level editing using volumetric representations. Instruct-NeRF2NeRF~\cite{haque2023instruct} allows users to edit a NeRF scene through text prompts, but such edits affect the entire scene rather than individual objects.
Our work addresses a different setting. Given an image, we isolate an arbitrary object, lift it into a 3D representation that supports editing, and then reinsert the modified object into the original scene. To restore occluded regions, we leverage powerful modern inpainting models~\cite{suvorov2022resolution}, allowing the final composite to remain coherent and visually plausible. This object-centric, compositional approach enables geometry-aware editing of in-the-wild images without requiring category-specific priors or multi-view input.

\section{Method}

The pipeline of our method, as illustrated in Fig.~\ref{fig:pipeline}, comprises the following steps: instance segmentation, 2D-to-3D generation, 3D editing, and inpainting the rendered edits of 3D shapes back into the image.

\subsection{Lifting 2D Instances to 3D Space}

Unlike existing image-based editing techniques that manipulate pixels in 2D space using generative models such as GANs or Diffusion, our approach focuses on editing 2D instances by crafting them as 3D objects. This method provides a more intuitive and human-aligned understanding of the physical world. The process begins by identifying the instances to be edited and lifting them into 3D space as a 3D representation.

Aiming at this goal, given an image containing instances to edit, we first employ SAM~\cite{kirillov2023segment} to isolate each individual. The resulting segments are then cropped to center these instances. These cropped images are subsequently processed using the off-the-shelf 3D generation model TRELLIS~\cite{xiang2025structured}, which produces the 3D Gaussian Splatting (3DGS)~\cite{kerbl3Dgaussians} representation, chosen for its fast rendering speed and high visual quality.

The obtained 3DGS is represented as a set of Gaussian kernels ${\mathcal{G}_i : (\mu_i, o_i, s_i, q_i, c_i)}$, where $\mu_i \in \mathbb{R}^3$ denotes the center of the Gaussian, $o_i \in \mathbb{R}^+$ represents the opacity, $s_i \in \mathbb{R}^{+3}$ corresponds to the scaling factors along the 3D axes, and $q_i \in \mathbb{R}^4$ is the quaternion representing the $SO(3)$ rotation of the Gaussian.

\subsection{Physics-Sensible Interactive Editing}
With the 3DGS representation ${\mathcal{G}_i : (\mu_i, o_i, s_i, q_i, c_i)}$ in hand, the next step is to efficiently edit it while ensuring adherence to physically plausible constraints. To achieve this, we adopt the concept of ARAP~\cite{sorkine2007rigid,huang2024sc}, enabling interactive editing of 3DGS while maintaining the physical prior that each local part of the 3D shape preserves as much rigidity as possible. This approach applied to 3DGS ensures that the edited results undergo plausible deformations while strictly adhering to user-defined constraints.

\begin{figure}[h]
\begin{center}  
\includegraphics[width=\linewidth]{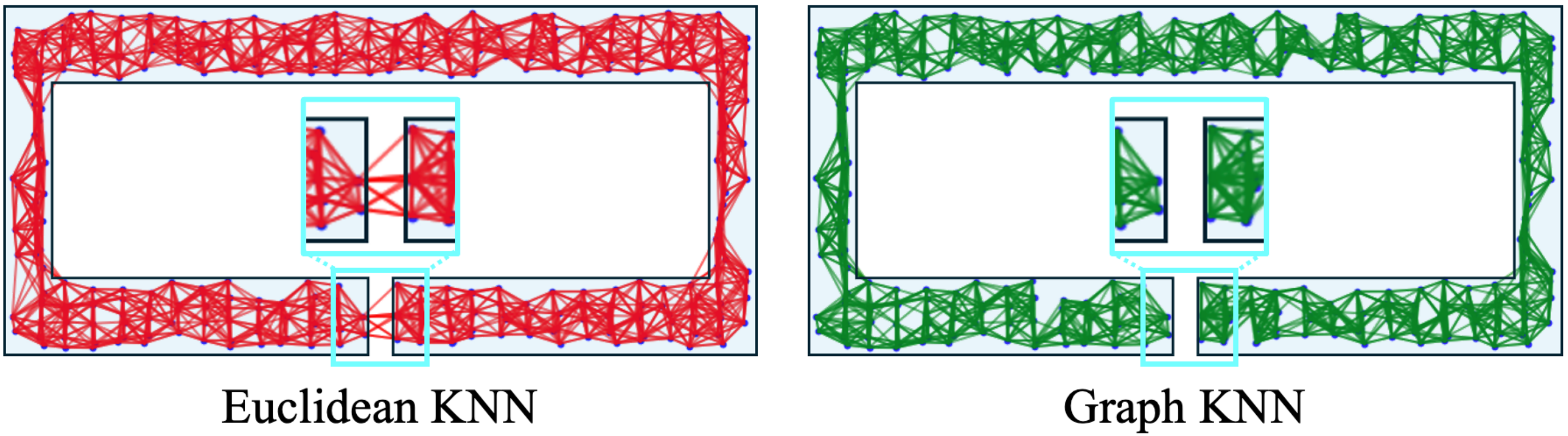}
\vspace{-10mm}
\end{center}
\caption{Graph better preserves topology than the Euclidean.}
\label{fig:graph}
\vspace{-4mm}
\end{figure}

Since 3DGS contains millions of Gaussian primitives to accurately approximate the target object, performing deformation based on per-Gaussian analysis is computationally prohibitive for real-time interaction and memory efficiency. To address this, we derive a set of coarse control points evenly distributed over the 3D shape using farthest-point sampling on the Gaussians. During our experiments, the number of control points is set to 512. Denoting control point positions before deformation as $p_i$ and their deformed positions as $p'_i$, we define the local rigidity energy as:

\begin{equation}
E(p') = \sum\limits_i \sum\limits_{j\in \mathcal{N}i} w_{ij} | (p'_i - p'_j) - R_i (p_i - p_j) |^2,
\label{eq:arap_energy}
\end{equation}
where $R_i$ is the estimated rotation of the local cell centered at $p_i$, $\mathcal{N}i$ is the neighborhood of $p_i$ consisting of $K=8$ nearest neighbors, and $w{ij}$ represents the importance weight of $p_j$ relative to $p_i$. This energy term is both translation- and rotation-invariant, penalizing any local non-rigid deformation.
Rather than directly applying $K$-nearest neighbors (K-NN) in Euclidean space, we first use $\frac{K}{2}$-nearest neighbors to connect control nodes and construct an initial graph. Then, we compute the $K$-NN based on the shortest path distances within this graph. As illustrated in Fig.~\ref{fig:graph}, this approach preserves the topology of the final connected graph by preventing control nodes from linking to overly distant points from separate regions.

When the user interacts with the system by clicking on the image and dragging $H$ selected control points to reposition them in 3D space, we impose strict constraints on these handle points:

\begin{equation}
p'_{h_i} = \tilde{p}_{h_i}, \quad i \in [H],
\end{equation}
where $h_i$ denotes the indices of the selected control points.

The term in Eq.~\ref{eq:arap_energy} can be interpreted as penalizing deviations in relative positions, centralized by the Laplacian matrix and rotated by the estimated local rotation. This can be expressed as:

\begin{equation}
Lp' = b,
\label{eq:laplacian}
\end{equation}
where $b = [\sum\limits_{j \in \mathcal{N}i} \frac{w_{ij}}{2} (R_i + R_j)(p_i - p_j)]^T$ is a constant vector, and $L$ is the Laplacian matrix, which centralizes positions relative to their weighted neighboring centers. The minimization of Eq.~\ref{eq:laplacian} can be solved linearly using the inverse or pseudo-inverse of $L$, depending on the constraint setup. The rows and columns corresponding to ${h_i}$ in $L$ and the rows in $b$ are removed to solve for the unconstrained points, while the constrained points remain fixed at user-defined positions.

The local rotation $R_i$ is estimated using Singular Value Decomposition (SVD) on the matrix~\cite{sorkine2017least}:
\begin{equation}
S_i = \sum\limits_{j\in\mathcal{N}i} w_{ij}(p_j - p_i)^T(p'_j - p'_i).
\end{equation}
By performing SVD on $S_i = U_i\Sigma_iV_i^T$, the local rotation $R_i$ is computed as $R_i = V_iU_i^T$.

The entire solving process is performed iteratively by alternating between optimizing $p'$ and updating ${R_i}$, gradually converging toward a global optimum~\cite{moon1996expectation}. In practice, three iterations are sufficient to achieve satisfactory results.

Finally, the deformation of Gaussians is driven by their neighboring control points using a Linear Blend Skinning (LBS)~\cite{magnenat1989joint} approach. We express the deformation of Gaussian parameters as:
\begin{equation}
    \begin{aligned}
        \mu'_i &= \sum\limits_{j\in \tilde{\mathcal{N}}_i} \tilde{w}_{i,j}\left(R_j \left(\mu_i-p_j \right) + p'_j \right), \\
        q'_i  &= \sum\limits_{j\in \tilde{\mathcal{N}}_i} \tilde{w}_{i,j} \text{Quat}\left( R_j \cdot \text{Rot}(q_i)  \right).
    \end{aligned}
\end{equation}
Here, $\tilde{\mathcal{N}}_i$ represents the $\tilde{K}=3$ nearest control points to the $i$-th Gaussian, and $\tilde{w}_{ij}$ are weights that decay with distance. This method ensures smooth, physically plausible deformations of the Gaussian representation while respecting user-defined constraints.

\subsection{3D Object Placing and Inpainting}

The final crucial step in our pipeline is to seamlessly reintegrate the edited 3D object back into the 2D image. The drop stage involves two concurrent processes: restoring the background occluded by the original object and rendering the edited object for final composition. The goal is to produce a coherent and photorealistic image of the manipulated instance placed in the original image.

Lifting the target object from the image inevitably leaves a "hole" in the background where the object was previously located. To create a plausible scene for re-integration, this occluded area must be semantically restored. We accomplish this using the initial segmentation mask generated by SAM. This mask precisely defines the region requiring inpainting.
We then employ a state-of-the-art resolution-robust inpainting network to fill the designated area. The network takes the masked image as input and synthesizes the missing background content, ensuring contextual and structural consistency with the surrounding pixels. This process results in a complete and clean background plate, ready to receive the edited object.

Concurrently, the manipulated 3D Gaussian Splatting(3DGS) model is rendered back into a 2D image from a consistent viewpoint. The final composition begins with a standard alpha blending of the rendered object onto the inpainted background. To eliminate any unnatural seams at the object's boundary, we perform a targeted inpainting-based refinement. A dilated mask is generated around the object's contour to isolate the transition area. This masked region is then inpainted using prompts like "seamless integration" to guide the model. This technique effectively harmonizes the boundary by synthesizing a contextually aware transition, ensuring that the final image is photorealistic and coherent. We utilize PixelHacker~\cite{xu2025pixelhacker} for inpainting the composed images, given its robust and exceptional performance.

% ------------------------------
% GAN-generated Images (Two 3x3 groups side-by-side)
% ------------------------------
\begin{figure*}[t]
\centering

% ------------ Left GAN image group ------------
\begin{minipage}[t]{0.48\textwidth}
    \centering
    \hspace{1.4em}
    \makebox[0.28\linewidth]{\centering \textbf{Input Image}}\hspace{0.03\linewidth}%
    \makebox[0.28\linewidth]{\centering \textbf{Edit Guidance}}\hspace{0.03\linewidth}%
    \makebox[0.28\linewidth]{\centering \textbf{Result}}%
    \vspace{0.03\linewidth}

    % Row 1
    \raisebox{-0.1em}{\rotatebox{90}{\small DragDiffusion}}\hspace{1em}%
    \includegraphics[width=0.28\linewidth]{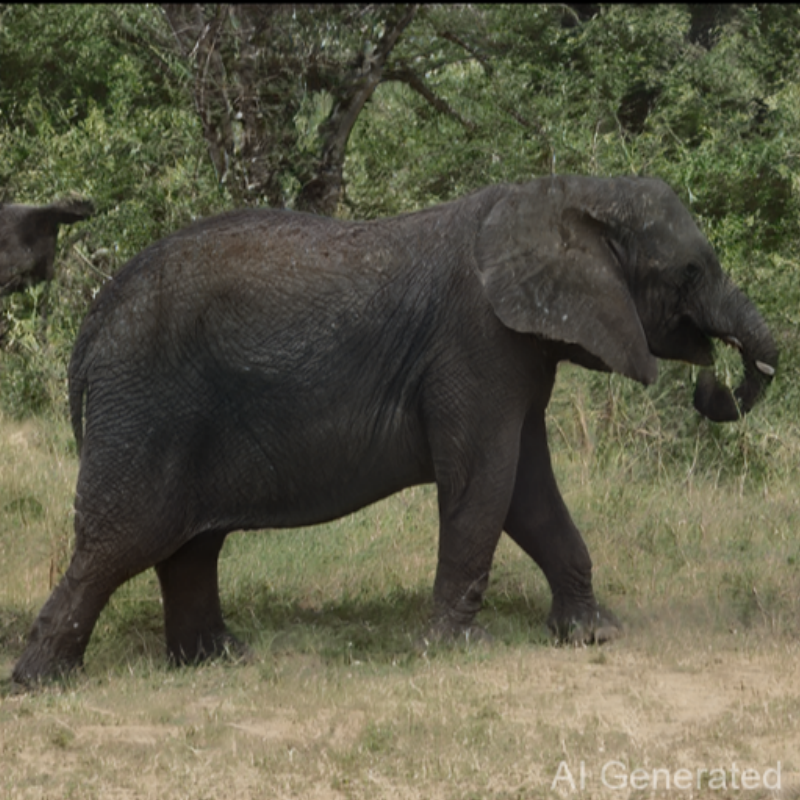}\hspace{0.03\linewidth}%
    \includegraphics[width=0.28\linewidth]{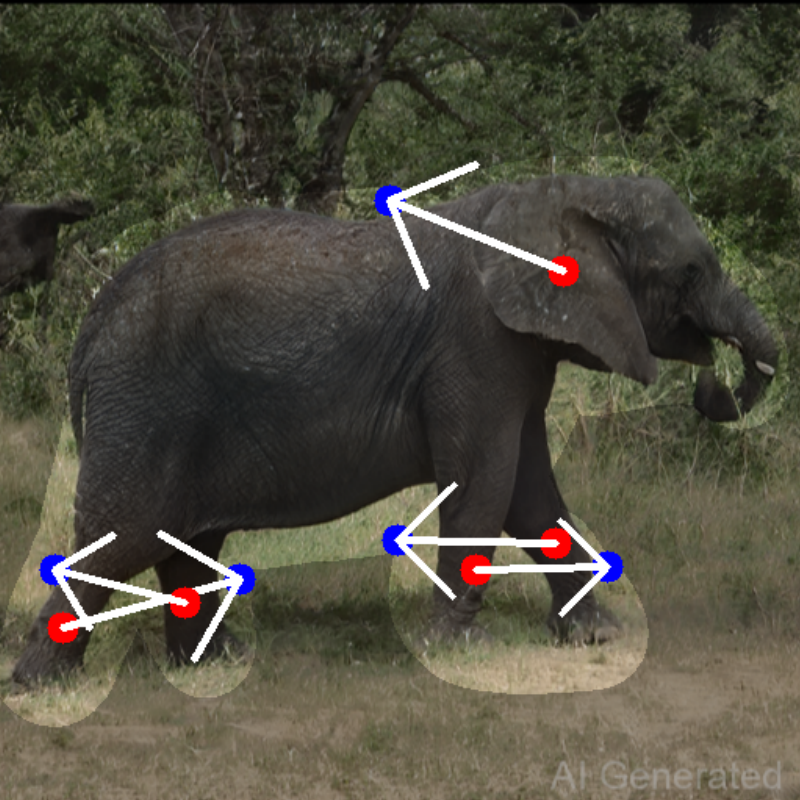}\hspace{0.03\linewidth}%
    \includegraphics[width=0.28\linewidth]{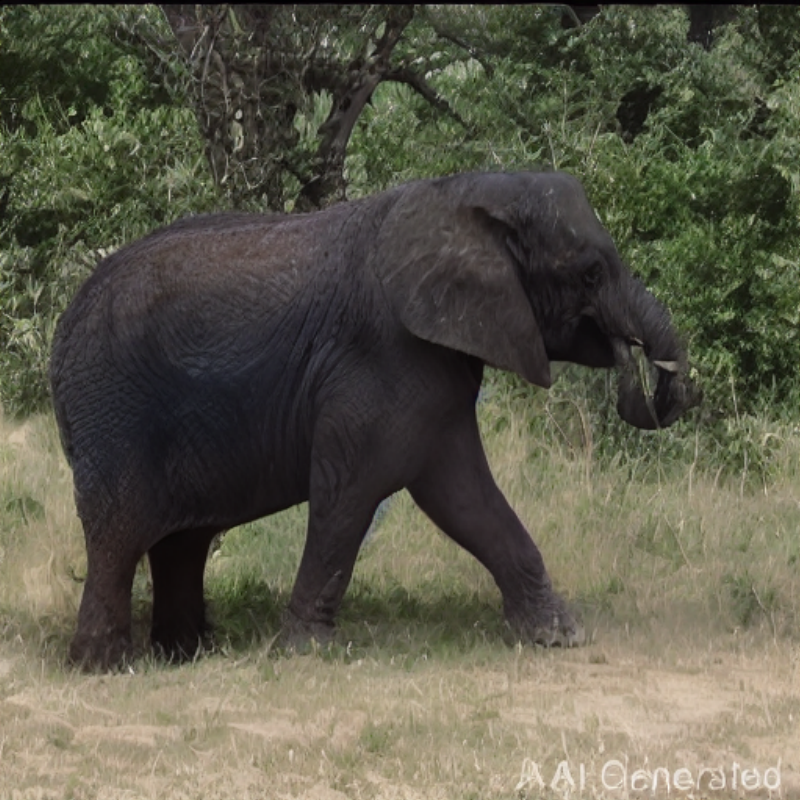} \\
    \vspace{0.03\linewidth}

    % Row 2
    \raisebox{0.5em}{\rotatebox{90}{\small DragGAN}}\hspace{1em}%
    \includegraphics[width=0.28\linewidth]{ori_elephant}\hspace{0.03\linewidth}%
    \includegraphics[width=0.28\linewidth]{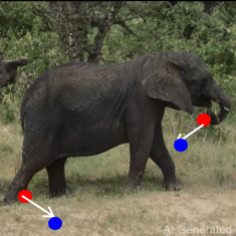}\hspace{0.03\linewidth}%
    \includegraphics[width=0.28\linewidth]{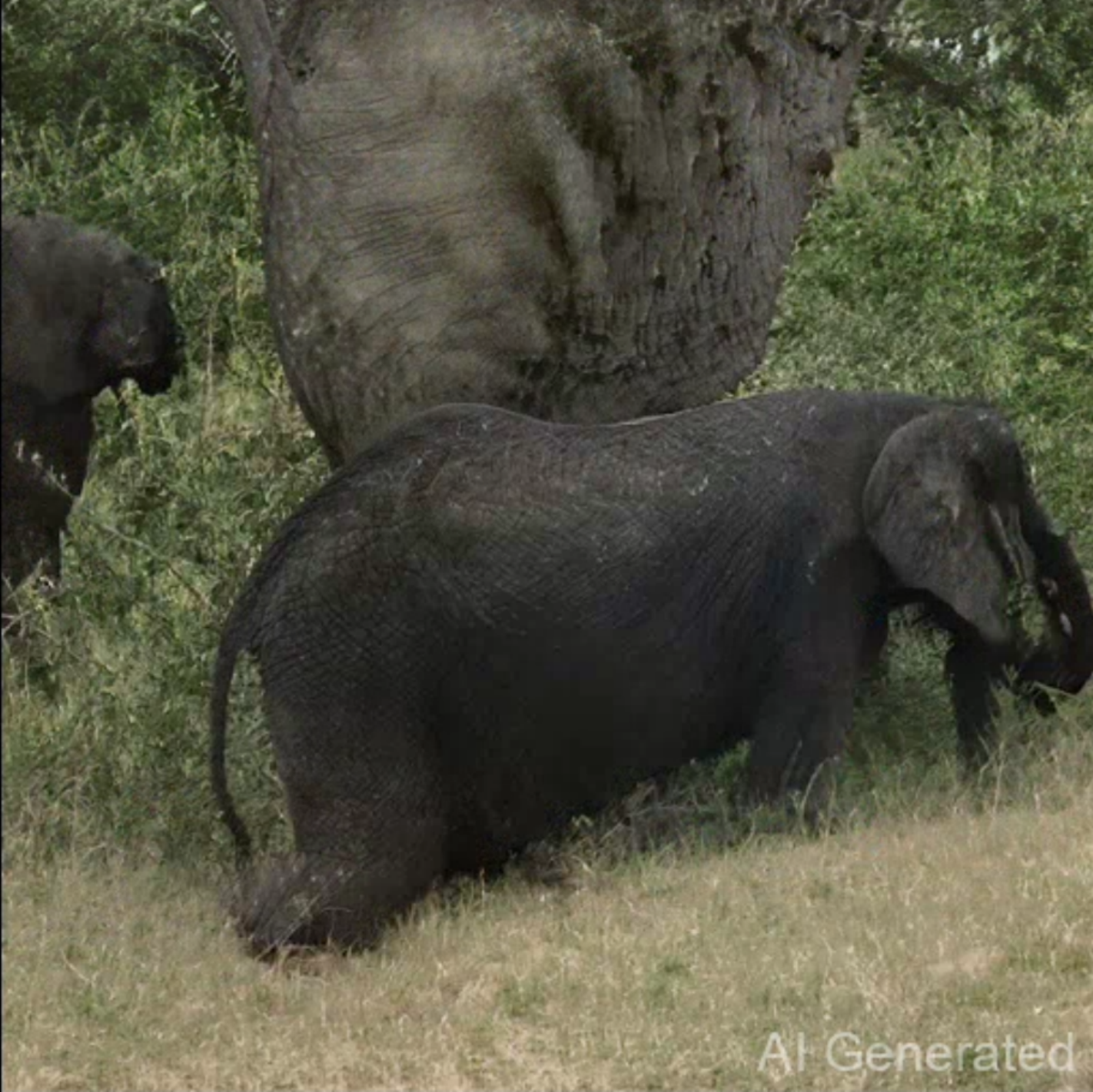} \\
    \vspace{0.03\linewidth}

    % Row 3
    \raisebox{0.7em}{\rotatebox{90}{\small Ours}}\hspace{1em}%
    \includegraphics[width=0.28\linewidth]{ori_elephant}\hspace{0.03\linewidth}%
    \includegraphics[width=0.28\linewidth]{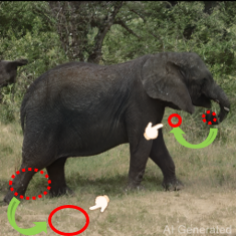}\hspace{0.03\linewidth}%
    \includegraphics[width=0.28\linewidth]{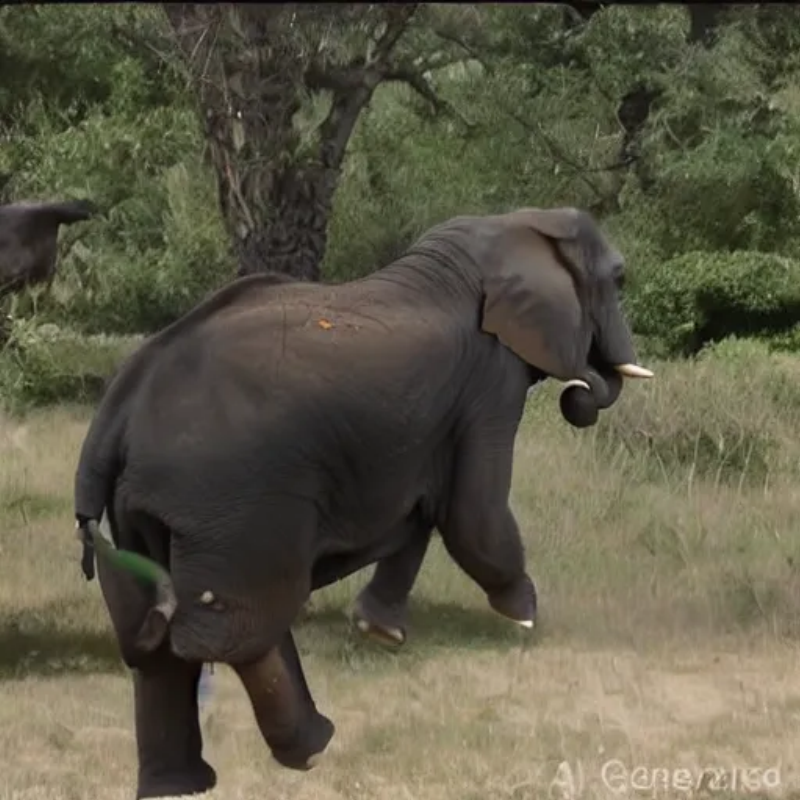}%
\end{minipage}
\hfill
% ------------ Right GAN image group ------------
\begin{minipage}[t]{0.48\textwidth}
    \centering
    \hspace{1.4em}
    \makebox[0.28\linewidth]{\centering \textbf{Input Image}}\hspace{0.03\linewidth}%
    \makebox[0.28\linewidth]{\centering \textbf{Edit Guidance}}\hspace{0.03\linewidth}%
    \makebox[0.28\linewidth]{\centering \textbf{Result}}%
    \vspace{0.03\linewidth}

    % Row 1
    \raisebox{0em}{\rotatebox{90}{\small DragDiffusion}}\hspace{1em}%
    \includegraphics[width=0.28\linewidth]{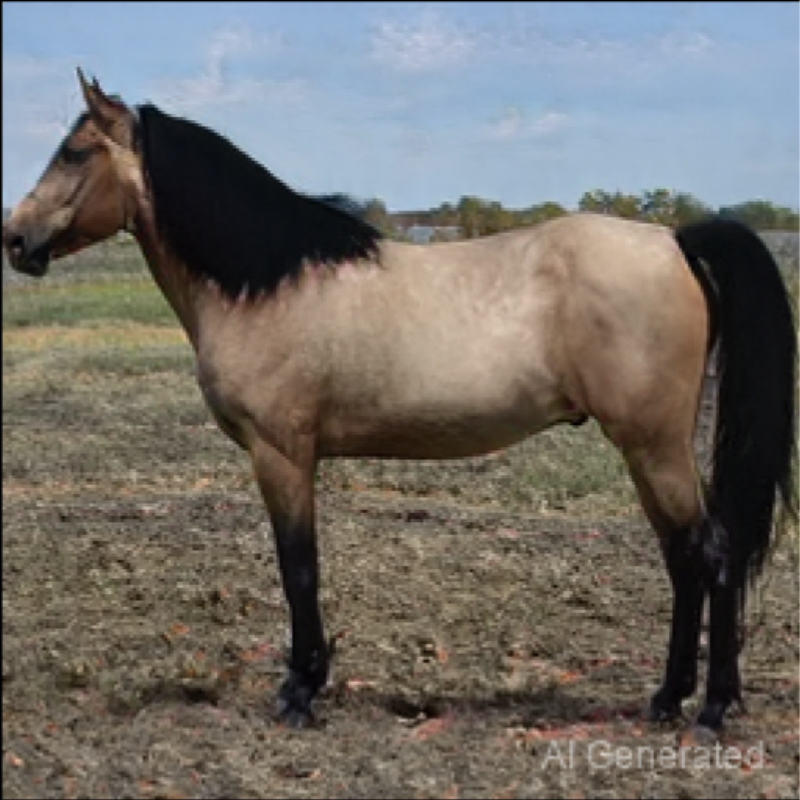}\hspace{0.03\linewidth}%
    \includegraphics[width=0.28\linewidth]{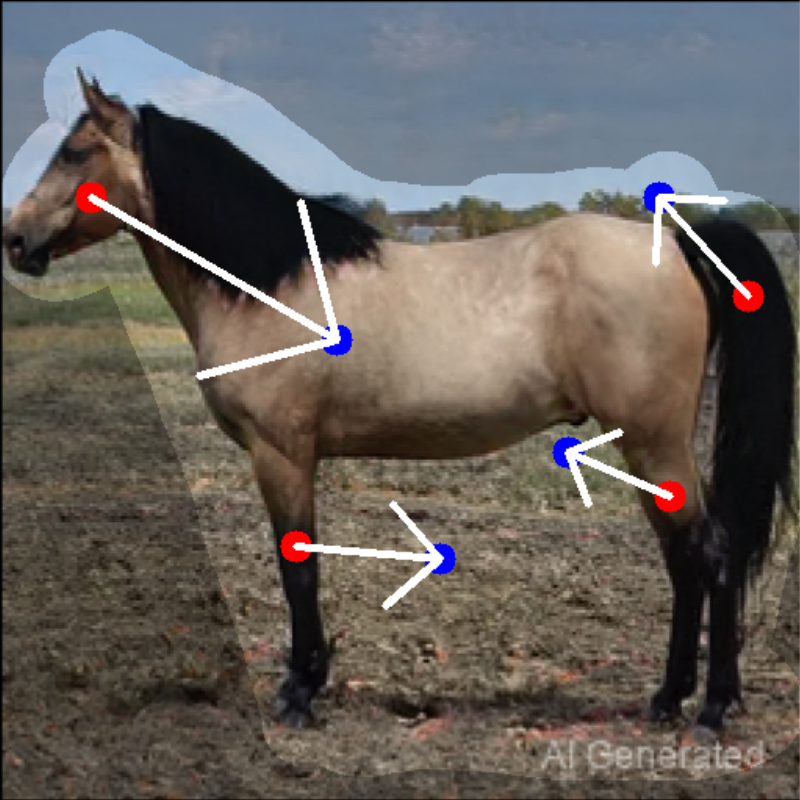}\hspace{0.03\linewidth}%
    \includegraphics[width=0.28\linewidth]{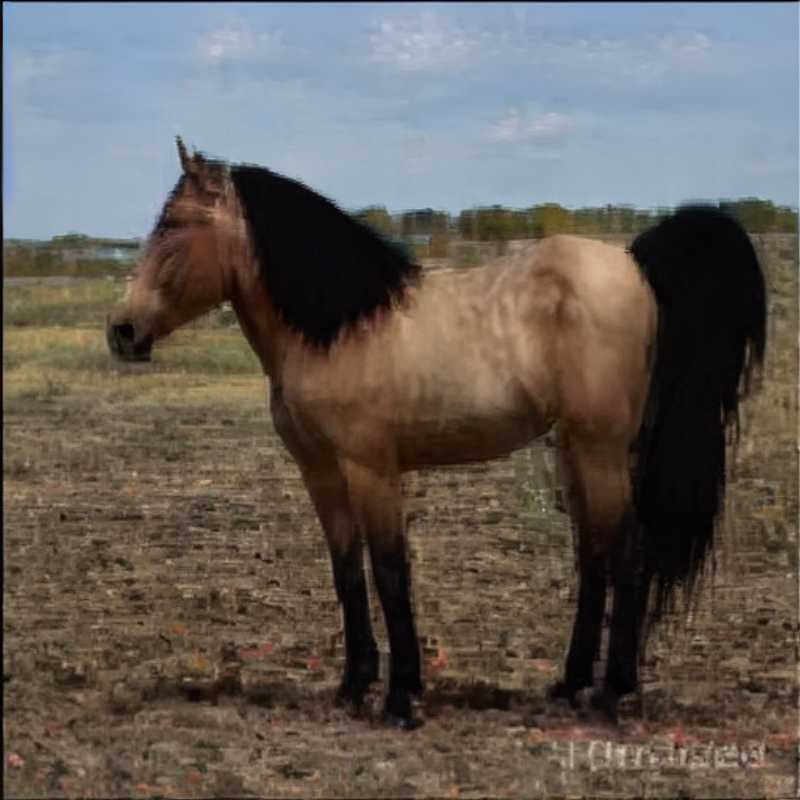} \\
    \vspace{0.03\linewidth}

    % Row 2
    \raisebox{0.8em}{\rotatebox{90}{\small DragGAN}}\hspace{1em}%
    \includegraphics[width=0.28\linewidth]{ori_horse.pdf}\hspace{0.03\linewidth}%
    \includegraphics[width=0.28\linewidth]{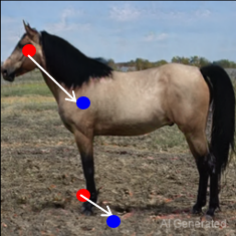}\hspace{0.03\linewidth}%
    \includegraphics[width=0.28\linewidth]{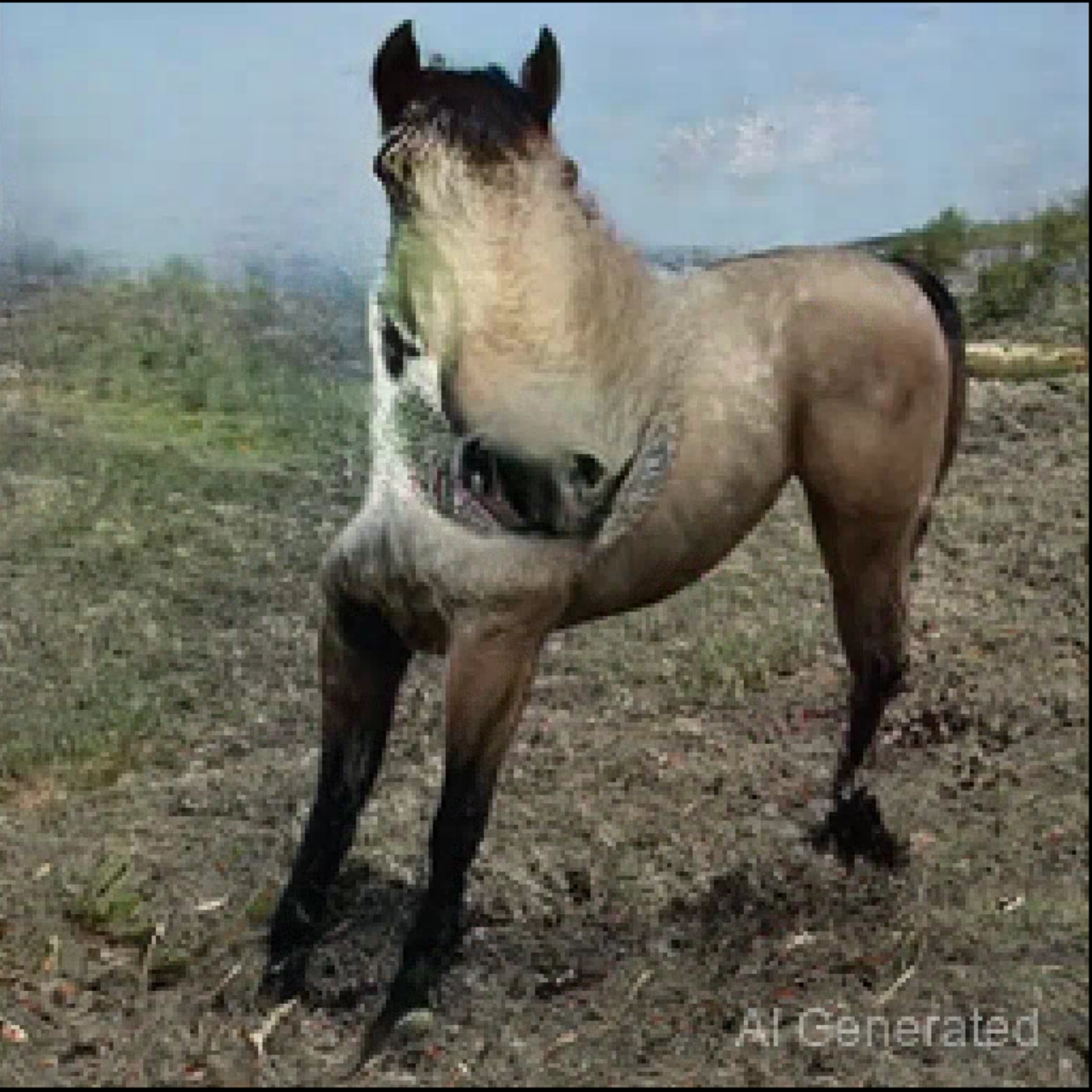} \\
    \vspace{0.03\linewidth}

    % Row 3
    \raisebox{1.1em}{\rotatebox{90}{\small Ours}}\hspace{1em}%
    \includegraphics[width=0.28\linewidth]{ori_horse.pdf}\hspace{0.03\linewidth}%
    \includegraphics[width=0.28\linewidth]{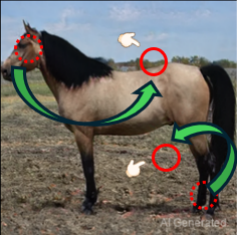}\hspace{0.03\linewidth}%
    \includegraphics[width=0.28\linewidth]{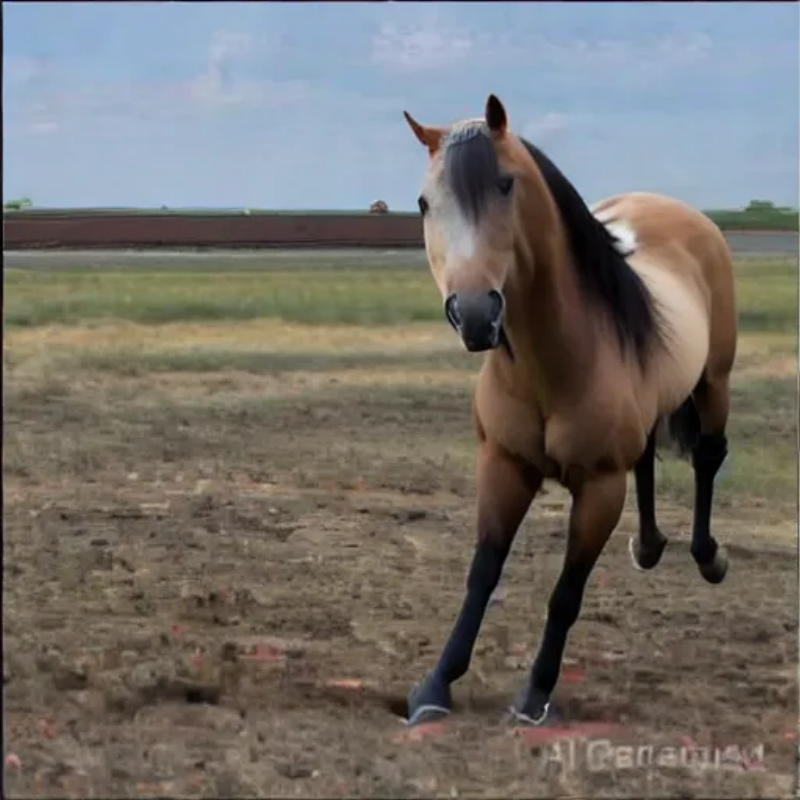}%
\end{minipage}

\caption{Qualitative comparisons of our method with DragDiffusion~\cite{shi2024dragdiffusion} and DragGAN~\cite{pan2023draggan} on natural images.}
\label{fig:gan_results}
\end{figure*}

% ------------------------------
% Real Images (Two 3x3 groups side-by-side)
% ------------------------------
\begin{figure*}[t]
\centering

% ------------ Left real image group ------------
\begin{minipage}[t]{0.48\textwidth}
    \centering
    \hspace{1.4em}
    \makebox[0.28\linewidth]{\centering \textbf{Input Image}}\hspace{0.03\linewidth}%
    \makebox[0.28\linewidth]{\centering \textbf{Edit Guidance}}\hspace{0.03\linewidth}%
    \makebox[0.28\linewidth]{\centering \textbf{Result}}%
    \vspace{0.03\linewidth}

    % Row 1
    \raisebox{0em}{\rotatebox{90}{\small DragDiffusion}}\hspace{1em}%
    \includegraphics[width=0.28\linewidth]{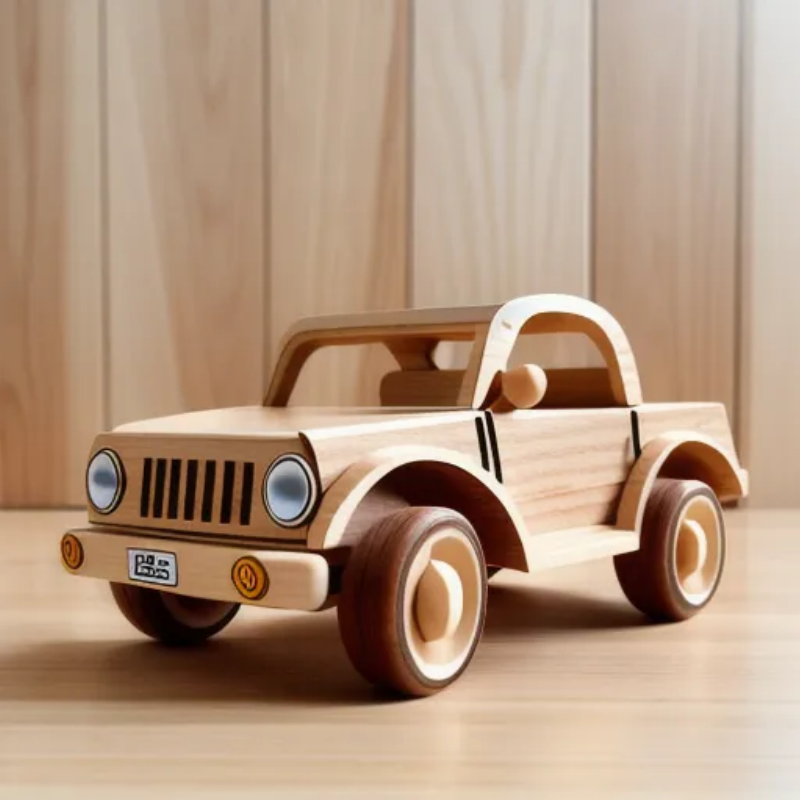}\hspace{0.03\linewidth}%
    \includegraphics[width=0.28\linewidth]{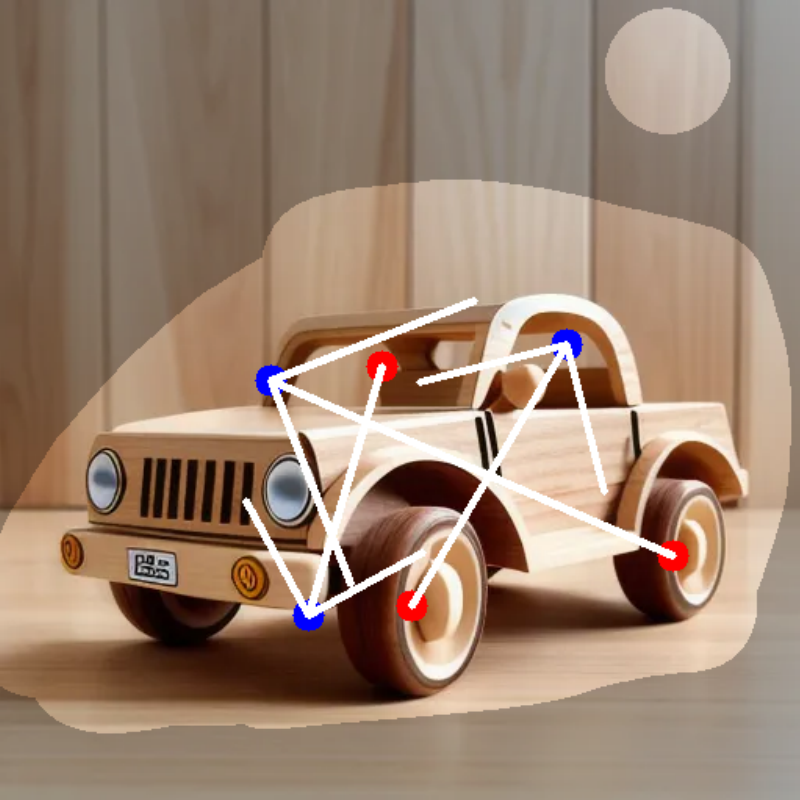}\hspace{0.03\linewidth}%
    \includegraphics[width=0.28\linewidth]{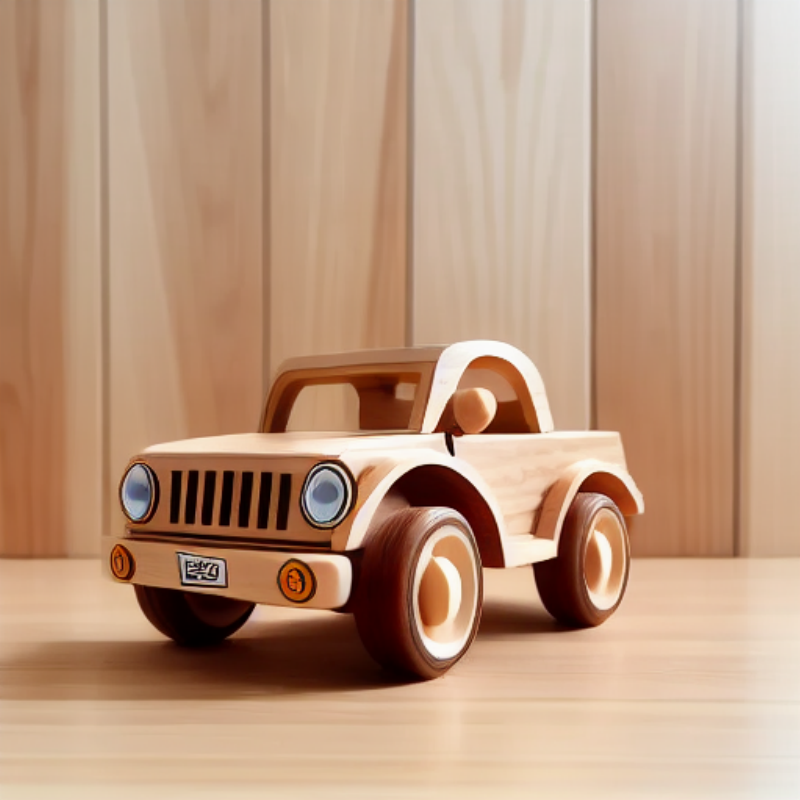} \\
    \vspace{0.03\linewidth}

    % % Row 2
    % \raisebox{0.8em}{\rotatebox{90}{\small DragGAN}}\hspace{1em}%
    % \includegraphics[width=0.28\linewidth]{3D-EDIT/images/our_results/brown_car.png}\hspace{0.03\linewidth}%
    % \includegraphics[width=0.28\linewidth]{3D-EDIT/images/compare/diffganours/pointsgan/na.png}\hspace{0.03\linewidth}%
    % \includegraphics[width=0.28\linewidth]{3D-EDIT/images/compare/diffganours/draggan/na.png} \\
    % \vspace{0.03\linewidth}

    % Row 3
    \raisebox{1.1em}{\rotatebox{90}{\small Ours}}\hspace{1em}%
    \includegraphics[width=0.28\linewidth]{ori_brown_car.pdf}\hspace{0.03\linewidth}%
    \includegraphics[width=0.28\linewidth]{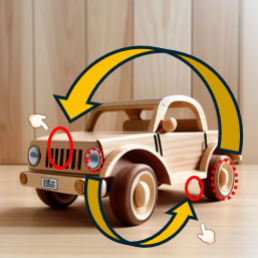}\hspace{0.03\linewidth}%
    \includegraphics[width=0.28\linewidth]{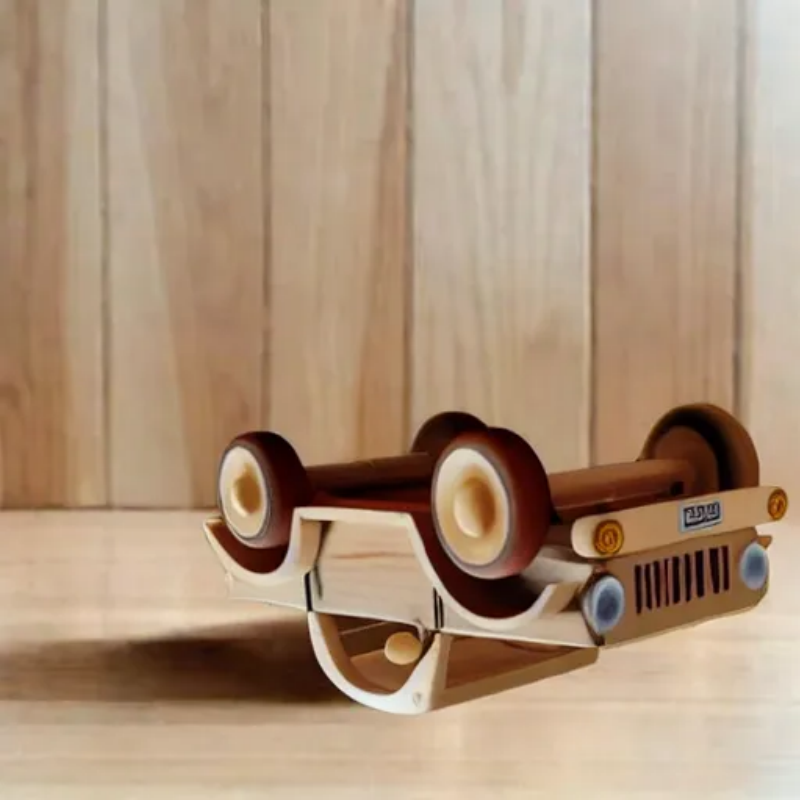}%
\end{minipage}
\hfill
% ------------ Right real image group ------------
\begin{minipage}[t]{0.48\textwidth}
    \centering
    \hspace{1.4em}
    \makebox[0.28\linewidth]{\centering \textbf{Input Image}}\hspace{0.03\linewidth}%
    \makebox[0.28\linewidth]{\centering \textbf{Edit Guidance}}\hspace{0.03\linewidth}%
    \makebox[0.28\linewidth]{\centering \textbf{Result}}%
    \vspace{0.03\linewidth}

    % Row 1
    \raisebox{0em}{\rotatebox{90}{\small DragDiffusion}}\hspace{1em}%
    \includegraphics[width=0.28\linewidth]{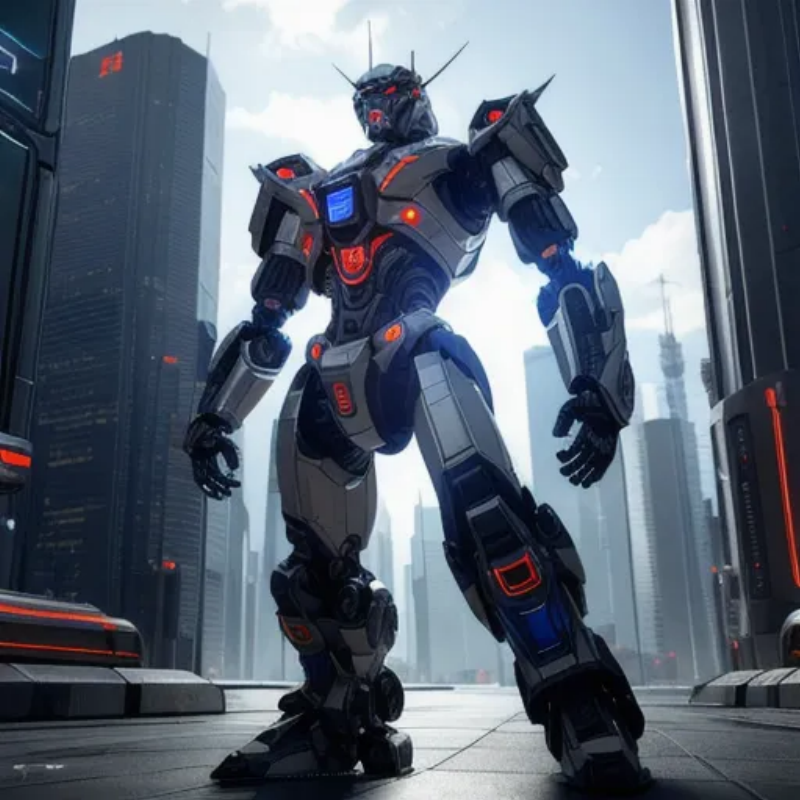}\hspace{0.03\linewidth}%
    \includegraphics[width=0.28\linewidth]{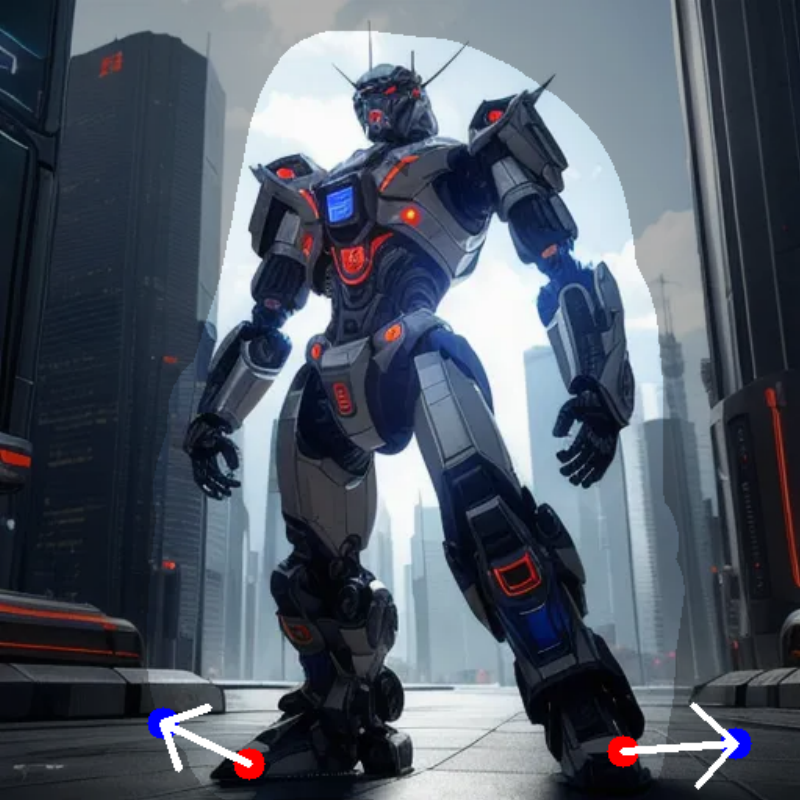}\hspace{0.03\linewidth}%
    \includegraphics[width=0.28\linewidth]{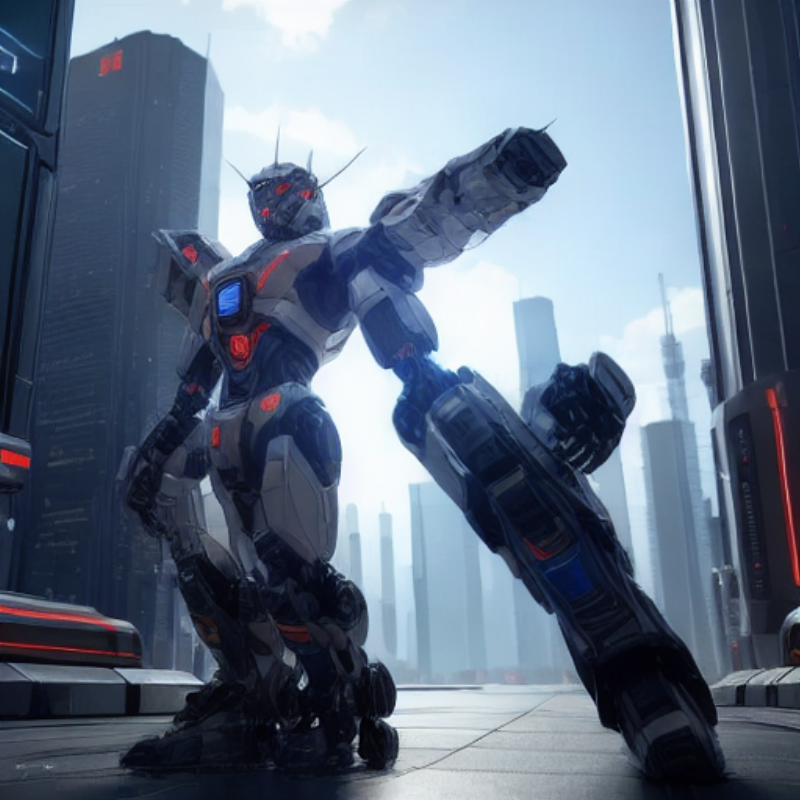} \\
    \vspace{0.03\linewidth}

    % % Row 2
    % \raisebox{0.8em}{\rotatebox{90}{\small DragGAN}}\hspace{1em}%
    % \includegraphics[width=0.28\linewidth]{3D-EDIT/images/our_results/mecha1.png}\hspace{0.03\linewidth}%
    % \includegraphics[width=0.28\linewidth]{3D-EDIT/images/compare/diffganours/pointsgan/na.png}\hspace{0.03\linewidth}%
    % \includegraphics[width=0.28\linewidth]{3D-EDIT/images/compare/diffganours/draggan/na.png} \\
    % \vspace{0.03\linewidth}

    % Row 3
    \raisebox{1.1em}{\rotatebox{90}{\small Ours}}\hspace{1em}%
    \includegraphics[width=0.28\linewidth]{ori_mecha1.pdf}\hspace{0.03\linewidth}%
    \includegraphics[width=0.28\linewidth]{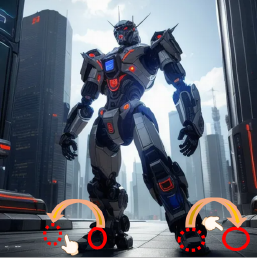}\hspace{0.03\linewidth}%
    \includegraphics[width=0.28\linewidth]{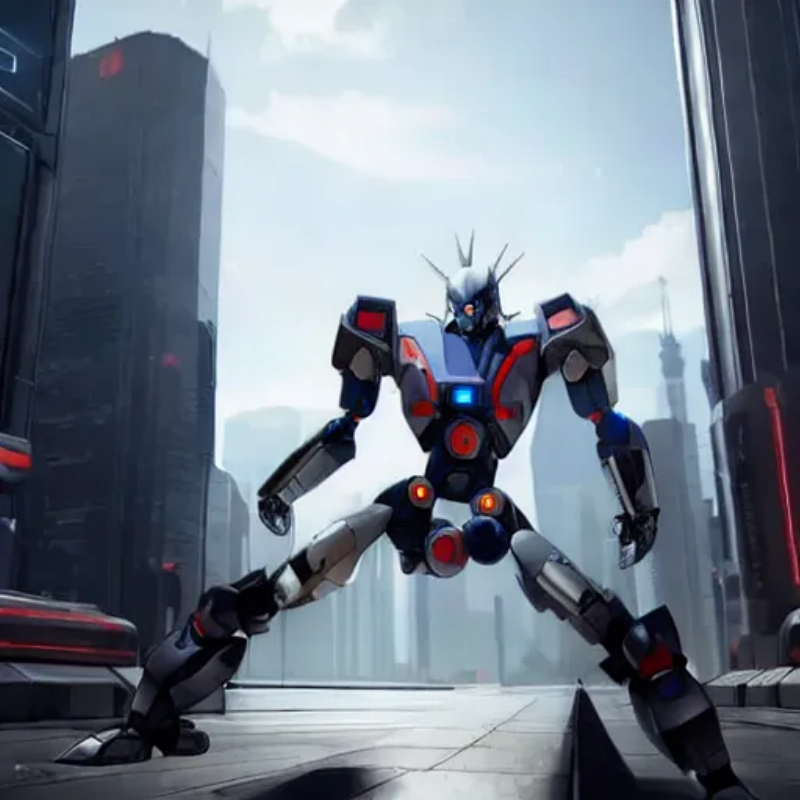}%
\end{minipage}

\caption{Qualitative comparisons of our method with DragDiffusion~\cite{shi2024dragdiffusion} on man-made objects. It is noteworthy that DragGAN~\cite{pan2023draggan} failed to produce any results for these input images.}
\label{fig:real_results}
\end{figure*}

\begin{figure}[htbp]
\centering
\begin{minipage}{0.48\textwidth}
    \centering
    % ---- Column Titles ----
    \makebox[0.3\linewidth]{\centering \textbf{Input Image}}%
    \makebox[0.3\linewidth]{\centering \textbf{Edit 1}}%
    \makebox[0.3\linewidth]{\centering \textbf{Edit 2}}%

    \vspace{0.5em}

    % Row 1
    \begin{subfigure}{0.3\linewidth}
        \includegraphics[width=\linewidth]{ori_brown_car.pdf}
    \end{subfigure}
    \begin{subfigure}{0.3\linewidth}
        \includegraphics[width=\linewidth]{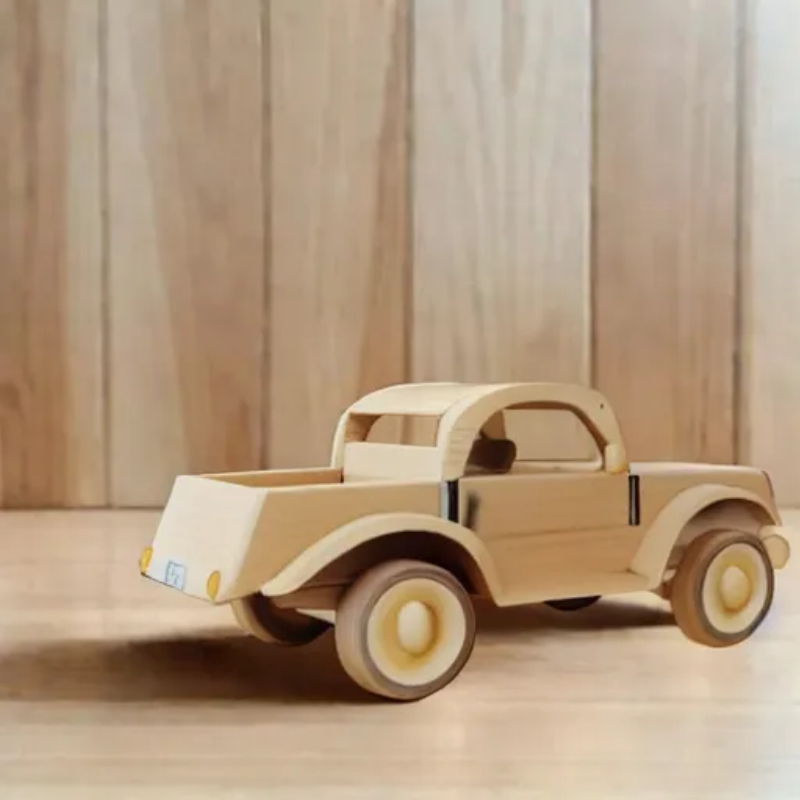}
    \end{subfigure}
    \begin{subfigure}{0.3\linewidth}
        \includegraphics[width=\linewidth]{our_results_brown_car_opposite.pdf}
    \end{subfigure}

    \vspace{0.5em}

    % Row 2
    \begin{subfigure}{0.3\linewidth}
        \includegraphics[width=\linewidth]{ori_mecha1.pdf}
    \end{subfigure}
    \begin{subfigure}{0.3\linewidth}
        \includegraphics[width=\linewidth]{our_results_mecha1_front.pdf}
    \end{subfigure}
    \begin{subfigure}{0.3\linewidth}
        \includegraphics[width=\linewidth]{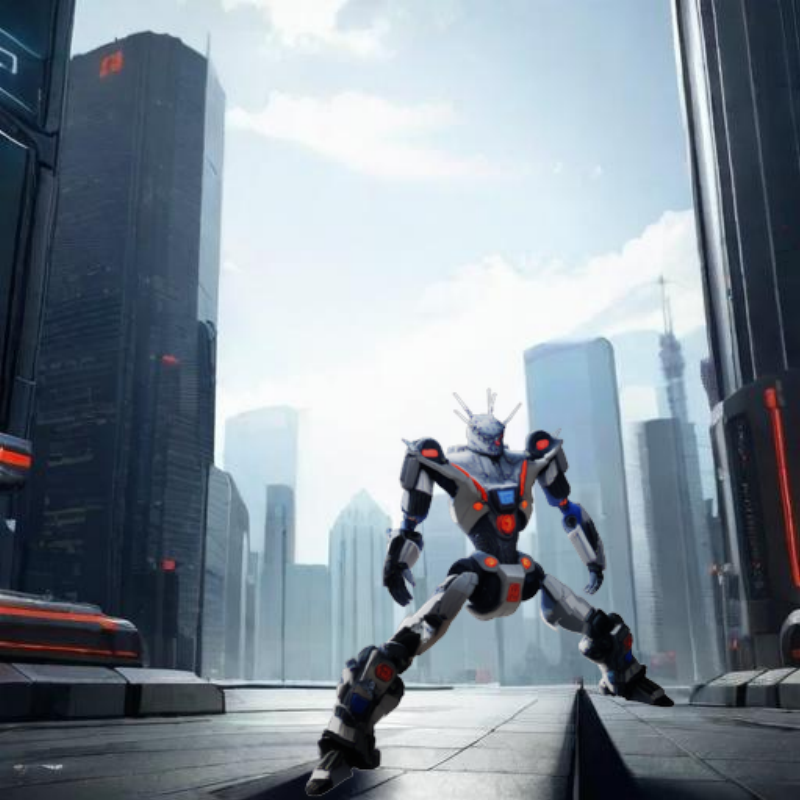}
    \end{subfigure}

    \vspace{0.5em}

    % Row 3
    \begin{subfigure}{0.3\linewidth}
        \includegraphics[width=\linewidth]{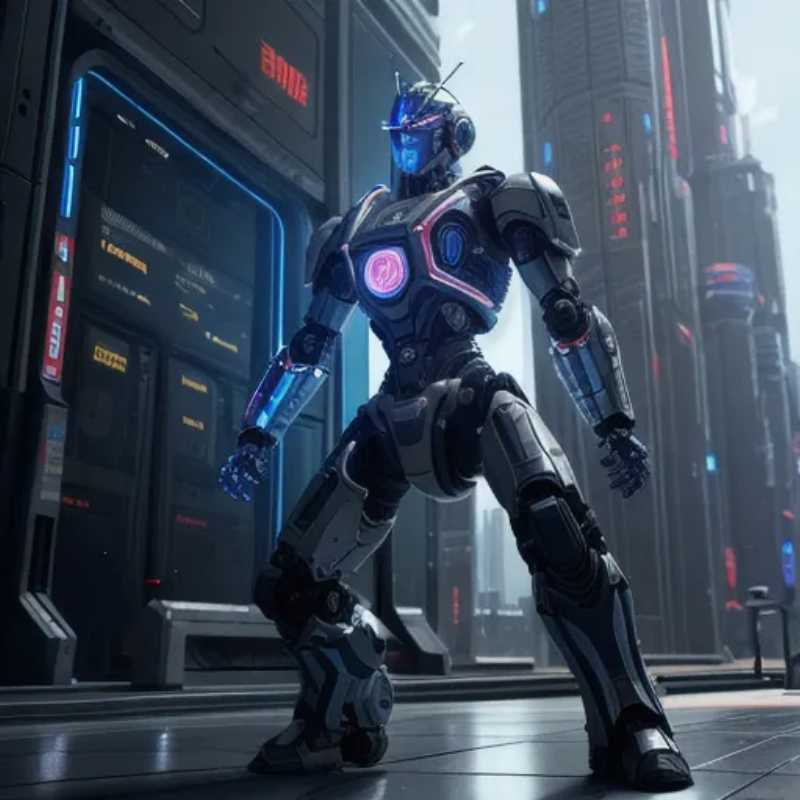}
    \end{subfigure}
    \begin{subfigure}{0.3\linewidth}
        \includegraphics[width=\linewidth]{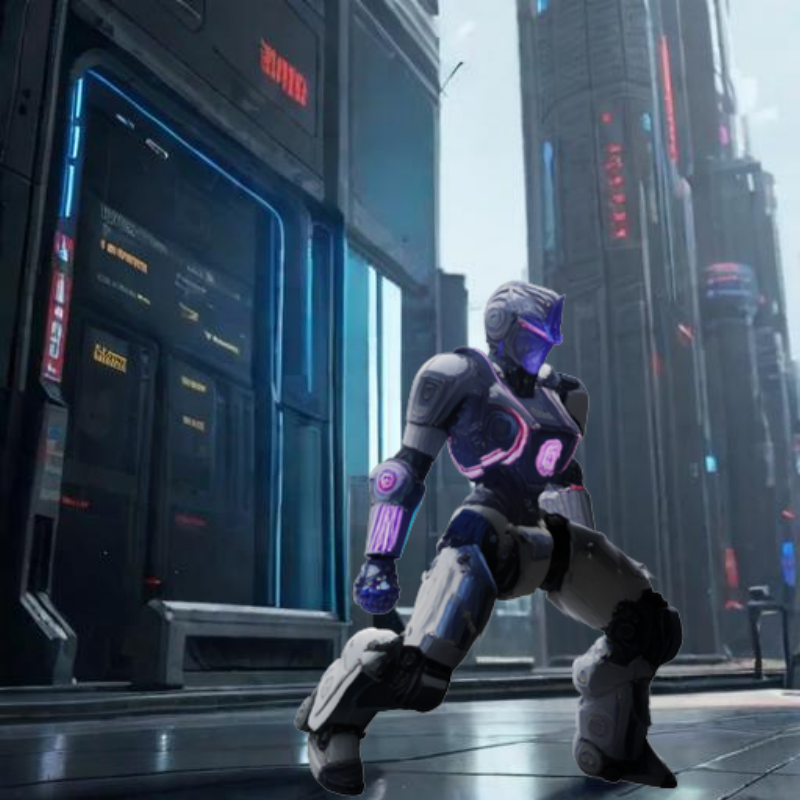}
    \end{subfigure}
    \begin{subfigure}{0.3\linewidth}
        \includegraphics[width=\linewidth]{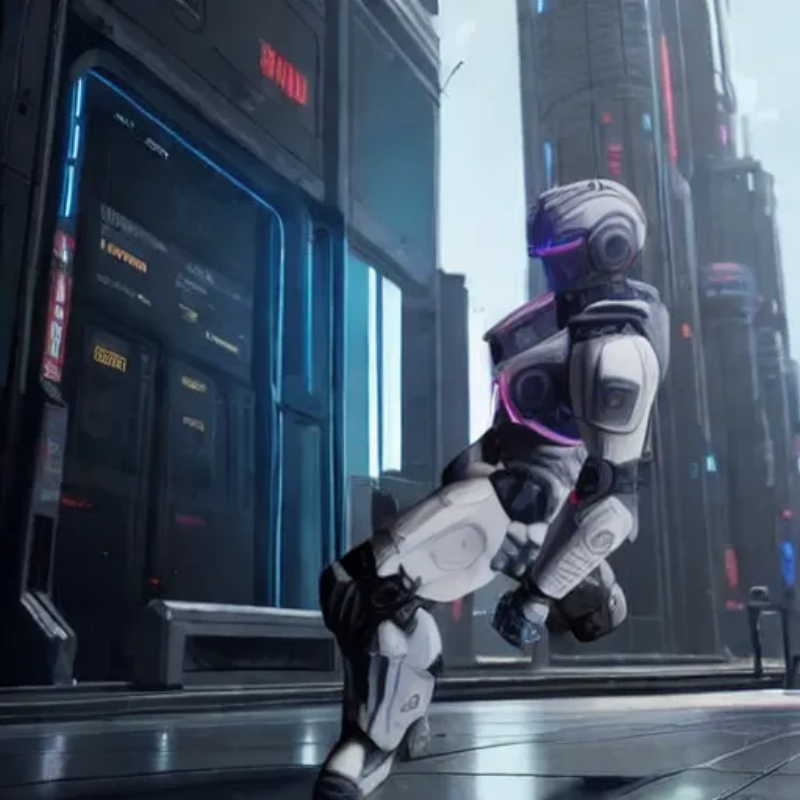}
    \end{subfigure}

    \vspace{0.5em}

    % Row 4
    \begin{subfigure}{0.3\linewidth}
        \includegraphics[width=\linewidth]{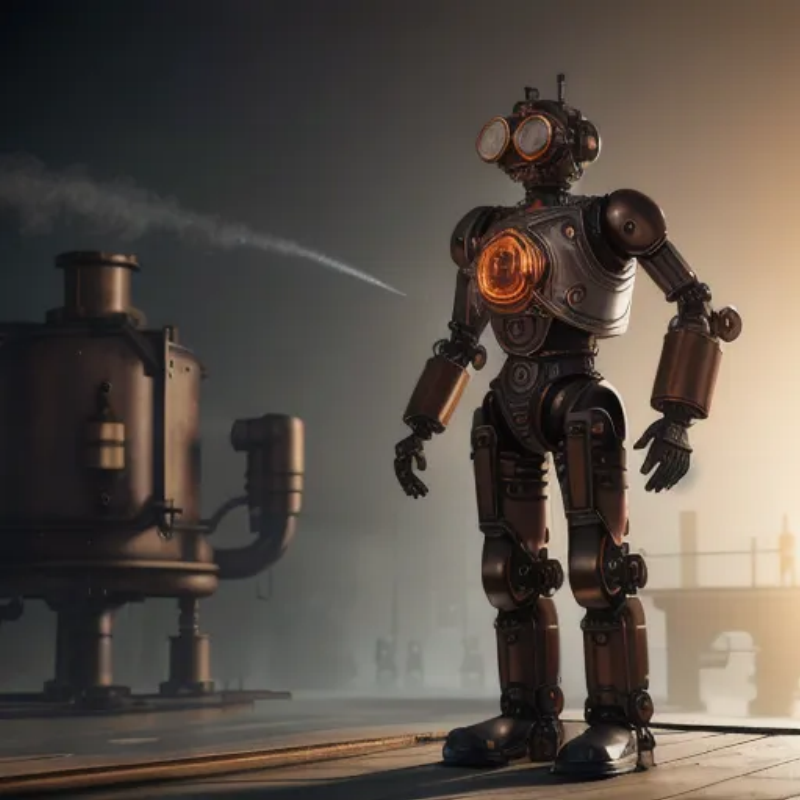}
    \end{subfigure}
    \begin{subfigure}{0.3\linewidth}
        \includegraphics[width=\linewidth]{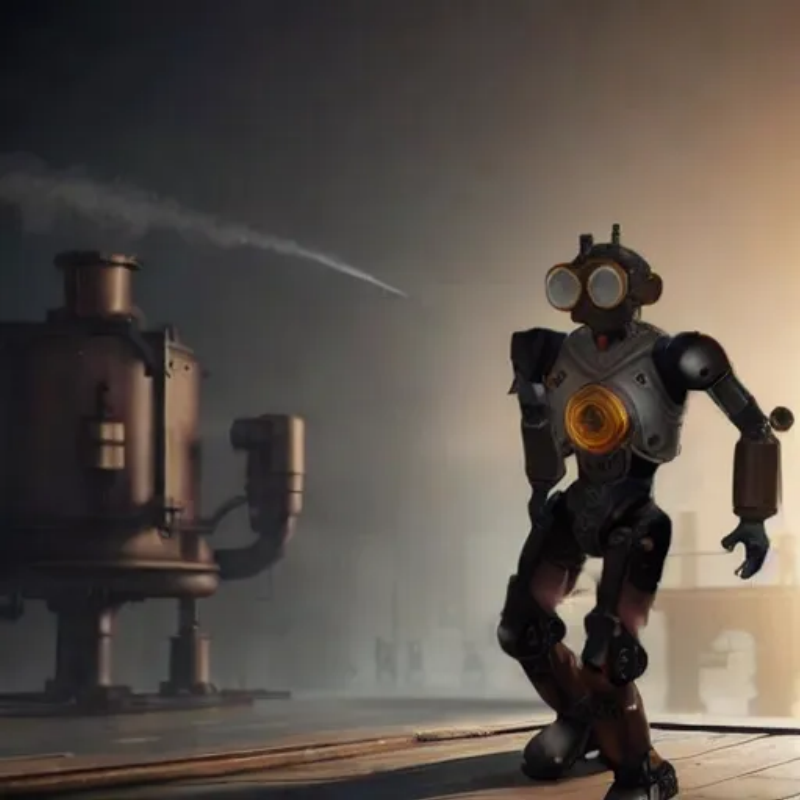}
    \end{subfigure}
    \begin{subfigure}{0.3\linewidth}
        \includegraphics[width=\linewidth]{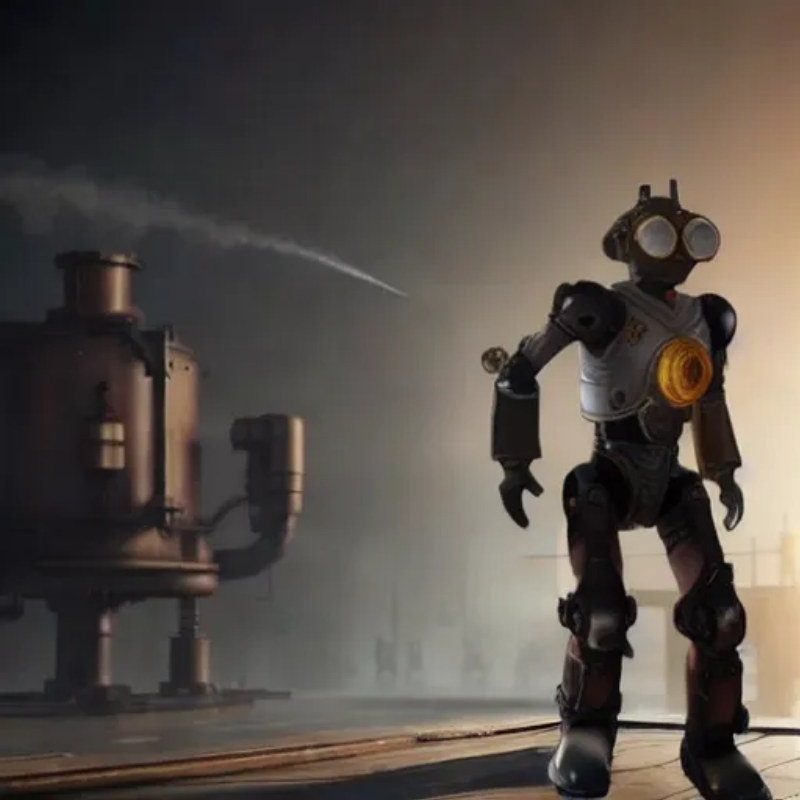}
    \end{subfigure}

    \vspace{0.5em}

    % Row 5
    \begin{subfigure}{0.3\linewidth}
        \includegraphics[width=\linewidth]{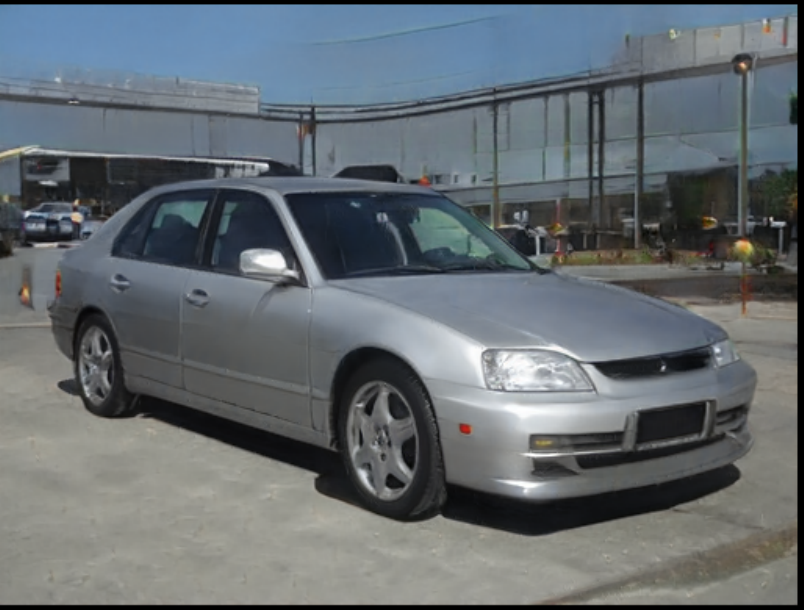}
    \end{subfigure}
    \begin{subfigure}{0.3\linewidth}
        \includegraphics[width=\linewidth]{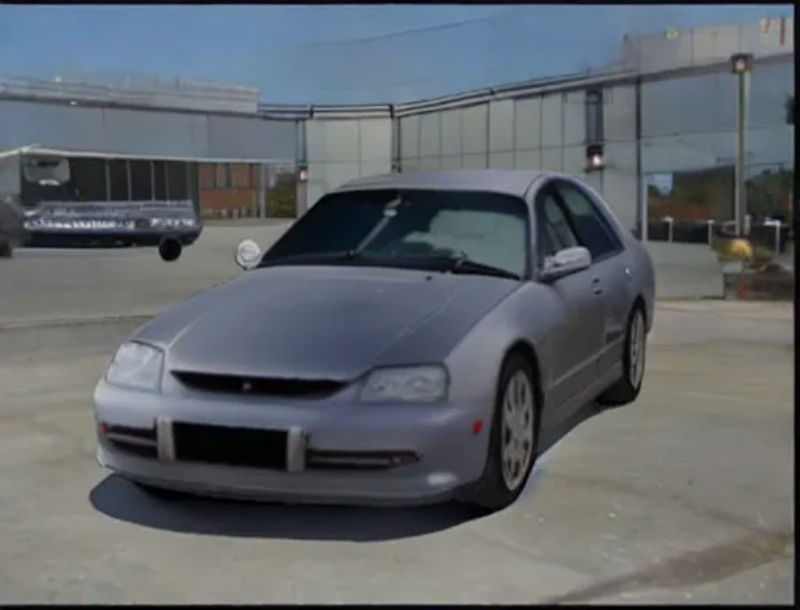}
    \end{subfigure}
    \begin{subfigure}{0.3\linewidth}
        \includegraphics[width=\linewidth]{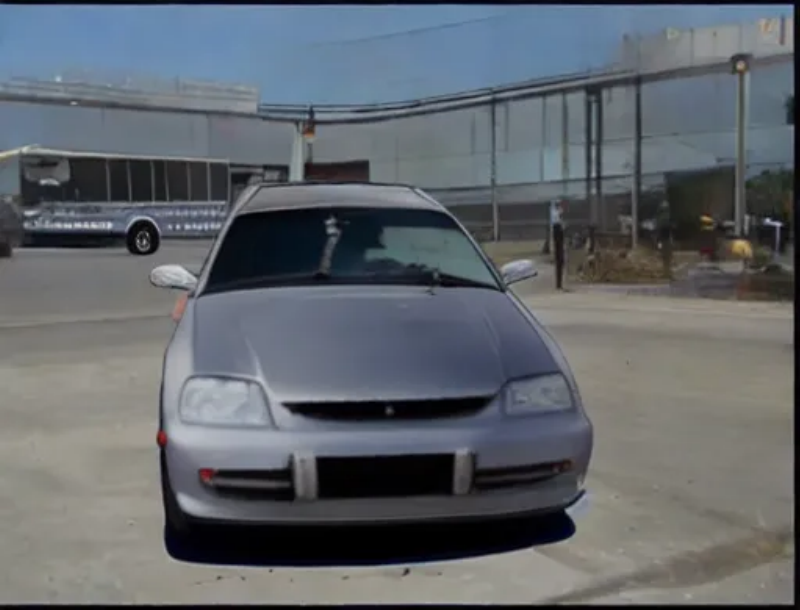}
    \end{subfigure}
\end{minipage}

\caption{Diverse editing results produced by our method.}
\label{fig:half_column_gallery}
\end{figure}

\section{Result}
\subsection{Qualitative Comparison}
We evaluated our proposed method against two state-of-the-art interactive editing approaches: DragDiffusion and DragGAN. These methods represent the current paradigm of point-based image manipulation, where users specify source and target positions to guide the editing process. Our comparison specifically focuses on editing scenarios that involve significant pose changes or viewpoint modifications. These transformations are particularly challenging for purely 2D-based approaches due to their inherent pixel-level manipulation, but they are naturally handled by our 3D-aware framework.

Our experiments were conducted on both natural images and images of man-made objects. 
As shown in Fig.\ref{fig:gan_results}, for natural images of specific targets such as elephants and horses, both DragDiffusion\cite{shi2024dragdiffusion} and DragGAN~\cite{pan2023draggan} produce edited results, but these are not strictly aligned with the given controls. For instance, the back legs and nose of the elephant cannot be moved forward and backward, respectively. DragDiffusion failed to rotate the horse's face toward the camera due to its in-plane nature. While DragGAN successfully rotated the horse's head, it produced unnatural results on the horse's face.

For images of more general man-made objects (Fig.~\ref{fig:real_results}), we observed that DragGAN failed to produce any results. Meanwhile, DragDiffusion was unable to achieve large rigid rotations of the target objects. Furthermore, DragDiffusion introduced semantic artifacts when handling significant non-rigid deformations, such as on the Gundam legs.

In contrast, our method consistently performs large-scale editing across a wide variety of images, producing high-fidelity outputs that robustly preserve object identity and scene consistency. This comprehensive evaluation demonstrates that our approach significantly outperforms the compared methods in challenging editing scenarios with flexible, non-rigid, accurate, and large-scale 3D control.

\subsection{Diverse Editing Results}
Our experiments extended to various other images, subjected to diverse editing manipulations. The showcased results in Figure \ref{fig:half_column_gallery} confirm that our method not only supports the large-scale edits users anticipate but also robustly preserves object identity. Qualitative evaluations highlight a significant reduction in visual artifacts and superior structural coherence compared to existing 2D methods, validating our 2D-3D-2D approach's efficacy.

\section{Conclusion}
Our work presents a real-time interactive 2D image editing method that performs manipulations on 2D instances in 3D space. This framework significantly outperforms prior interactive approaches, e.g, DragGAN and DragDiffusion, enabling large-scale edits across a wide range of images while consistently preserving object identity. However, the fidelity of our edits heavily depends on the quality of 3D reconstruction, which remains challenging for inputs with occlusions, severe lighting conditions, or distant object placement. Moreover, existing off-the-shelf inpainting methods fail to consistently produce satisfactory results. Using the composed images as a coarse draft, training a refinement model to specifically post-process these images is left as future work to achieve higher visual quality and better harmony between the instances and environments.

% {
%     \small
%     \bibliographystyle{ieeenat_fullname}
%     \bibliography{main}
% }

% WARNING: do not forget to delete the supplementary pages from your submission 
% \input{sec/X_suppl}

\end{document}